\documentclass[runningheads]{llncs}
\usepackage{graphicx}
\usepackage{amsmath}
\usepackage{amssymb}
\usepackage{booktabs}
\usepackage{subfigure}
\usepackage{xcolor}

\PassOptionsToPackage{hyphens}{url}\usepackage[hidelinks]{hyperref} 

\begin{document}
%
\title{A Comparison of Recent Algorithms for Symbolic Regression to Genetic Programming}
%

%
\author{Yousef A. Radwan \and
Gabriel Kronberger\orcidID{0000-0002-3012-3189} \and
Stephan Winkler\orcidID{0000-0002-5196-4294}}
\authorrunning{Y. A. Radwan et al.}
\titlerunning{A Comparison of Recent Algorithms for SR to GP}
%
\institute{University of Applied Sciences Upper Austria, Heuristic and Evolutionary Algorithms Laboratory, Softwarepark 11, 4232 Hagenberg \\
\email{yo.radwan@nu.edu.eg}\\
\email{\{gabriel.kronberger,stephan.winkler\}@fh-hagenberg.at}\\
}
\maketitle              
\begin{abstract}
Symbolic regression is a machine learning method with the goal to produce interpretable results. Unlike other machine learning methods such as, e.g. random forests or neural networks, which are opaque, symbolic regression aims to model and map data in a way that can be understood by scientists. Recent advancements,  have attempted to bridge the gap between these two fields; new methodologies attempt to fuse the mapping power of neural networks and deep learning techniques with the explanatory power of symbolic regression. In this paper, we examine these new emerging systems and test the performance of an end-to-end transformer model for symbolic regression versus the reigning traditional methods based on genetic programming that have spearheaded symbolic regression throughout the years. We compare these systems on novel datasets to avoid bias to older methods who were improved on well-known benchmark datasets. Our results show that traditional GP methods as implemented e.g., by Operon still remain superior to two recently published symbolic regression methods.

\end{abstract}
\section{Motivation}


Symbolic regression (SR)~\cite{Koza1992,Kronberger2024} is a supervised machine learning task where a functional mapping between one or multiple independent variables and usually one dependent variable has to be identified based on a dataset containing observations of the independent and dependent variables. In contrast to most other regression methods, the goal is to find a mathematical expression or formula for the regression function, whereby both, the expression structure, and fitting parameter values must be found by the algorithm. Thus, symbolic regression provides the potential to find short interpretable expressions based only on a set of observational data \cite{Bomarito2021,Makke2024}. In this context, SR has been called a hypothesis generation method \cite{keren2023computational,Koza1992}.

Our main goal is to gain a better understanding of recently developed symbolic regression (SR) methods, which mainly use approaches based on deep learning of neural networks to produce interpretable models. 
Even though the results published in papers are often impressive, the authors often do not invest the time to compare with established algorithms using, e.g., SRBench \cite{la2021contemporary}, which is a collection of SR problem instances and a well-curated and maintained list of SR implementations that can be used for benchmarking purposes. Since such comparisons are missing, there is a lack of understanding how well these methods work in practice and whether they truly improve upon established approaches.
Even if the methods are tested on SRBench, there is the danger of optimizing methods for the benchmark (Goodhart's law \cite{Strathern1997}) instead of trying to improve methods in general. This can lead to the effect that the benchmark becomes less useful over time.

For our experiments we have therefore selected a list of real-world datasets from different domains of engineering which are so far not widely used in SR publications and most importantly are not yet contained in SRBench. This set of problem instances can be understood as a validation set that allows us to detect if algorithms are overfit on the SRBench problem instances.


\section{Literature Review}

The need to find mathematical models that describe observations and can be used for predictions, exists since the beginning of all scientific endeavors. Since researchers first tackled this with physics and mathematics-inspired methods, to the use of evolutionary algorithms such as genetic programming, to the entry of the newest contender which is machine learning. Although machine learning, and particularly deep learning, has struggled to enter this area in the past due to its black-box nature, recent creative methodologies have leveraged the power of neural networks and newer architectures such as transformers to create interpretable models that can qualify under the symbolic regression (SR) umbrella. Affenzeller et~al.~\cite{affenzeller2020eurocast} compared traditional machine learning methods including neural networks to GP-based symbolic regression and found that GP-based SR can produce more accurate and more interpretable models models.

SR methods are generally split into regression-based methods (linear and non-linear), expression tree-based (genetic programming, reinforcement learning, transformers), physics-inspired (i.e. AI-Feynman), and math-inspired (i.e., symbolic-metamodel) \cite{Makke2024}.


\subsection{Genetic Programming Methods}
Genetic programming \cite{Koza1992} was the main system for symbolic regression for a considerable portion of its history. It is built upon the idea of generating a population of models and then iteratively improving the population using a methodology similar to the idea of natural selection where weak models are pruned out and better models are selected to generate new, adapted models for the new population. It also includes the idea of mutations which add random changes to models during propagation.

In \cite{Vaddireddy_2020}, the authors discussed the use of Gene Expression Programming (GEP) and Sequential Threshold Ridge Regression (STRidge) algorithms in symbolic regression. These methods are used to extract hidden physics from sparse observational data. The effectiveness of these algorithms is demonstrated in various applications, including equation discovery and truncation error analysis, showcasing their ability to identify complex physics problems.

Furthermore, the authors of \cite{Vaddireddy_2020} emphasized the significance of feature selection and engineering in model discovery approaches and demonstrate the potential of these techniques in complex physics problems. The most significant difference between GEP and GP is the fixed vs variable length representations used in the algorithms.


\subsubsection{Integration of Domain Knowledge into Genetic Programming}
Conventional GP methods for symbolic regression generally only used prediction error as their guide through the search space. However, with small datasets or when the data samples do not sufficiently cover the input space, prediction error does not serve as high-quality guidance \cite{Kubal_k_2020}. This leads these methods to generate partly incorrect models that exhibit incorrect steady-state characteristics or local behavior \cite{Kubal_k_2020,KUBALIK2021115210} or that do not generalize well to points outside of the training set. Multiple papers have tackled the challenge of incorporating domain knowledge to bridge this gap. Some approached this by the addition of discrete data samples on which candidate models are checked, serving as a sort of internal representation of a constraint \cite{Kubal_k_2020}. Other papers proposed a multi-objective symbolic regression framework \cite{KUBALIK2021115210} that optimizes models with respect to prediction error and also their compliance with desired physical properties. This framework also proposed a method for selecting a single final model out of the pool of candidate output models.

Another approach is shape-con\-strain\-ed symbolic regression. Kronberger et~al. \cite{Kronberger_2022} investigate an approach by adding constraints that target the function image and its derivatives. This allows the user to enforce the monotonicity of the function over a selected input range. As a side effect domain knowledge can be used to improve extrapolation accuracy.

\subsubsection{Reducing the Size of the Search Space}
Some methods have extended GP with extra steps to reduce the size of the search space by removing algebraically isomorphic expressions and limiting the complexity of expressions.
Exhaustive Symbolic Regression \cite{ESR} and Grammar Enumeration \cite{GE} rely on heuristics-guided exhaustive search of the function space to find all possible structures and 
then evaluation on a specified cost function to find the best fit. 

Another example of a SR system which limits the search space is the Inter\-action-Trans\-forma\-tion Evolutionary Algorithm \cite{defranca2020interactiontransformation} which can represent functions as interactions between predictors and the application of a single transformation function. Other papers have made use of this representation with other systems such as multi-layer neural networks with the weights being adjusted following the Extreme Learning Machine procedure \cite{DEFRANCA2021609}. This improved or maintained performance while reducing computational cost.


\subsection{Physics-inspired Symbolic Regression}
As a short overview of physics-inspired methods for symbolic regression, many methodologies have been formulated using physics theorems and inspirations. One example is the QLattice system~\cite{broløs2021approach} which is inspired by Richard Feynman's path integral formulation. This method explores many potential SR models and formulates them as graphs that can be interpreted as mathematical equations. This gives the user fairly strong control over the models' interpretability, complexity and performance. Unfortunately, the QLattice code has not been published. We therefore cannot use it for our own experiments.

Another SR method which can be characterized as physics-inspired is the AI-Feynman system \cite{udrescu2020ai}, which is available as Python code. The performance of AI Feynman relative to other SR implementations is already well understood as it is also included within the SRBench project. The enhanced version of AI-Feynman \cite{udrescu2020ai2} incorporates Pareto-optimality, aiming for the best accuracy relative to complexity. This version shows marked improvement in noise and data robustness, surpassing previous methods in formula discovery. It introduces a technique to identify generalized symmetries using neural network gradients and extends symbolic regression to probability distributions using normalizing flows and statistical hypothesis testing, enhancing the search process's efficiency and robustness.


The Scientist-Machine Equation Detector (SciMED) \cite{keren2023computational} merges the expertise of scientific disciplines with cutting-edge symbolic regression techniques in a scientist-in-the-loop approach. It combines a genetic programming-based wrapper selection method with automated machine learning and two-tiered symbolic regression strategies. 
%
AI-Descartes ~\cite{Cornelio2023} is another system that allows to integrate physics-based domain knowledge into symbolic regression using mixed integer non-linear programming.

\subsection{Neural Network-Based \& Deep Learning Methods}
With the entrance of deep learning methods into the symbolic regression field, there have been numerous contributions and systems that have relied on neural network based architectures~\cite{Kim_2021,petersen2021deep}. The main feature of these architectures is their ability to be trained end-to-end and their better performance in extrapolation and prediction for points outside the training set as seen in \cite{Kim_2021}. The main drawback that restricted deep learning methods was their natural complexity limiting interpretability but Kim et al.~\cite{Kim_2021} and many other recent papers have proposed solutions to this issue.

Li et al.~\cite{li2023metasymnet} proposed a novel neural network that could dynamically adjust its structure in real time, allowing for both expansion and contraction. They mainly addressed the issue that the fixed network architecture often gave rise to redundancy in network structure and parameters and sought to remove this restriction. Their use of adaptive neural networks and innovative activation functions such as the PANGU meta-function allowed them to evolve the trained neural network into a usable mathematical expression.

\subsubsection{Sequence-to-Sequence Models \& Transformers}
More recent deep learning contributions have made use of the popular transformer architecture or sequence to sequence architectures. With different variations from convolutional models such as \cite{biggio2020seq2seq} to recurrent network-based approaches such as  \cite{d2022deep} and \cite{dAscoli2024odeformer}, the symbolic regression research community has recently been flooded with transformer-based solutions and sequence-to-sequence models.

Their ability to map numerical data to corresponding symbolic equations and their encode-decoder structure make these architectures attractive for the SR task. Early papers used deep learning architectures to generate function structure and then optimize constants as a secondary step, while later papers attempted full function generation in one step \cite{vastl2022symformer,DeepSR}. 
They showed impressive extrapolation performance and high versatility. d'Ascoli et al. \cite{d2022deep} even proclaimed that their model outperformed Mathematica functions in sequence extrapolation and recurrence prediction and has highlighted the power of transformers in recurrent sequences in particular.

The major drawback is computational cost and training time, which can be on the order of days.
To address this, some papers have resorted to large scale pre-training \cite{biggio2021neural,dAscoli2024odeformer}. Thereby, the transformer is pretrained for the task of generating a symbolic equation from a set of input-output pairs. Thus, at test time, the model is just queried with a new set of points and the output is used to guide the search for the equation. This approach has shown to improve with the presence of more pretraining data and more compute.

\subsubsection{Deep Generative Networks}
Some papers have approached symbolic regression using  generative AI models. Using conditional generative models and large language models such as GPT, they have shown that this approach may also be viable, although not usually in one step and more of a good primer for function search.
Deep Generative Symbolic Regression (DGSR) \cite{holt2023deep} leverages pre-trained conditional generative models to encode equation invariances and provides a foundation for subsequent optimization steps. The paper demonstrates that DGSR achieves higher recovery rates of true equations, especially with a larger number of input variables, and is more computationally efficient at inference than state-of-the-art reinforcement learning solutions in symbolic regression. Holt et al. \cite{holt2023deep} highlight the advantages of DGSR, particularly in terms of scalability with the number of input variables and computational efficiency.

SymbolicGPT \cite{valipour2021symbolicgpt}, a new transformer-based language model for symbolic regression, leverages the strengths of large language models, including high performance and flexibility. Valipour et al. report impressive capabilities in accuracy, speed, and data efficiency, outperforming other models in these areas~\cite{valipour2021symbolicgpt}.

\subsection{Hybrid Methods}
Mundhenk et al. \cite{mundhenk2021symbolic} describe a hybrid method that combines neural-guided search with genetic programming for symbolic regression and other combinatorial optimization problems. The approach involves using a neural-guided component to generate initial populations for genetic programming, which evolves to yield progressively better starting points. They report, recovering 65\% more expressions from benchmark tasks, outperforming Deep Symbolic Regression~\cite{petersen2021deep}. 


In summary, many novel and reportedly powerful methods for symbolic regression have been proposed in the past few years fuel by the increased capabilities of deep learning architectures and models. However, the comparison with genetic programming based methods or classical machine learning methods is often lacking. We found that the code for many of the methods mentioned above is available online. However, with very few exceptions we failed to run the system on our own datasets. In many cases the code is simply dumped with minimal documentation to an online repository for the purpose of the publication and then abandoned. As a consequence we selected only two systems: \emph{end-to-end symbolic regression using transformers (E2E)} \cite{Kim_2021} and the \emph{Scientist-Machine Equation Detector (SciMED)} \cite{keren2023computational} for our comparison and used HeuristicLab~\cite{Wagner2014} and Operon~\cite{Burlacu2020} as representative implementations of tree-based genetic programming for symbolic regression.
SciMED provides different pipelines to generate models. The GA-SR pipeline uses the gplearn Python library\footnote{\url{https://gplearn.readthedocs.io/en/stable/}} for tree-based genetic programming for symbolic regression and the AutoML pipeline uses TPOT~\cite{TPOT}, which is a Python library that allows to optimize machine learning pipelines using genetic programming. 
Table~\ref{tab:systems} lists the software implementations that we have chosen for our experiment and their capabilities.

\begin{table}

\centering
     \caption{Characteristics and features of the software implementations selected for our experiments.}
     \label{tab:systems}
     \begin{tabular}{l|ccccc}
                     & HeuristicLab & Operon     &  E2E &  SciMED  &  SciMED \\
                    &               &            &         & AutoML   & GA-SR\\
   
        \hline
        Year         & 2014         & 2020       &  2022  &  2022  &  2022 \\
        Gen. prog.   & \checkmark   & \checkmark &   &  &  \checkmark  \\
        Neural net.  &              &            &  \checkmark & &  \\
        Dom. know.   & \checkmark   & & & \checkmark & \checkmark \\
     \end{tabular}

\end{table}

\section{Benchmarking Experiments}
In running our experiments, we sought to equalize the playing field among the models to allow as fair a comparison as possible given the vast differences between the models at hand. Each model had a different approach which will be detailed in the following. In most instances, if the accompanying paper offered default parameters or a model checkpoint to use, those parameters were used unless they incurred too much runtime as will be detailed more in the following sections.

In the case of E2E and SciMED the experiments were run on a RTX 2060 GPU with 6GB of VRAM. There are a few notable exceptions, which are mentioned in the corresponding segment, which were run on an M3 Pro chip, the performance differences are noted in case they are of interest.
The HeuristicLab and Operon experiments where run on an Intel Core i5-10400 CPU.

\subsection{Description of datasets}
We used datasets from \cite{Kronberger2024} with the same training and test splits as described in the book. The \emph{chemical 1 (tower)} and \emph{chemical 2 (competition)} datasets stem from continuous processes at Dow Chemical \cite{Kordon2008}. The \emph{tower} dataset was originally described in \cite{Vladislavleva2009}. The target variable is the propylene concentration measured at the top of a distillation tower and the input variables are process parameters. The \emph{competition} dataset has as target variable expensive but noisy lab data of the chemical composition (output) of the end-product, and 57 input variables with cheap process measurements, such as temperatures, pressures, and flows (inputs). This dataset was used in the symbolic regression competition at EvoStar conference 2010\footnote{\url{https://web.archive.org/web/20120628140646/http://casnew.iti.upv.es/index.php/evocompetitions/105-symregcompetition}}.

The \emph{friction} datasets were sponsored by Miba Frictec company and contain as target variable the friction coefficient for different types of friction materials as measured on an industrial friction testbench. The input variables are sliding velocity, pressure and temperature of friction materials. We used two separate datasets from the same set of experiments. In the first dataset, the target variable is the static coefficient of friction, and for the second dataset the target variable is the dynamic coefficient of friction. The dataset was originally described in \cite{Kronberger_GECCO_2018}, where the nominal variable for the friction material was included into the symbolic regression models using factor variables. In our experiments, we simply used a one-hot-encoding as this is supported by all tested software systems.

The \emph{flow stress} dataset was sponsored by LKR Light Metals Technologies, Austrian Institute of Technology, and contains measurements from dilatometer experiments using samples of a well-known aluminium alloy (AA6082). The target variable is the flow stress and the input variables are the strain the strain rate and the temperature. The dataset was originally described in \cite{Kabliman2021}, but we here only used a subset for constant strain rate of 0.1 to simplify the problem and speed up the experiments.

The \emph{battery} datasets were originally collected and published by the NASA Ames Research Center \cite{Saha2007} and are described in \cite{Goebel2008}. We used two datasets with preprocessed data from \cite{Kronberger2024} where the goal is to predict the remaining duration of discharge based on cell voltage, discharge current and cell temperature after ten minutes and twenty minutes of discharge under constant conditions.

Finally, the two \emph{Nikuradse} datasets contain measurements for the flow friction in rough pipes~\cite{Nikuradse1933}. Symbolic regression results for this dataset have been reported recently in \cite{Reichardt2020} and \cite{Kronberger_2024_ppsn}. We used two versions of the dataset: in the first dataset the target variable is the turbulent friction $\lambda$, and we use two input variables, the Reynolds number and the relative roughness $r/k$. The second dataset represents Prandtl's collapse, where the target is a transformed turbulent friction factor, and the input variable is a nonlinear combination of the relative roughness, and the Reynolds number \cite{Reichardt2020}. The datasets have been extracted directly from the tables in \cite{Nikuradse1933}.
All of the datasets were also split into training and test in the same manner for all our experiments. These splits can be seen below in Table \ref{train_test_table}:

\begin{table}[h]
\centering
\caption{Training and testing ranges for all datasets}
\begin{tabular}{lcc}
\hline
\textbf{Dataset}            & \textbf{Training partition} & \textbf{Testing partition} \\ \hline
Chemical 1 (tower)               & $0\ldots3135$                    & 3136\ldots4998                \\
Chemical 2 (Competition)         & $0\ldots710$                     & 711\ldots1065                 \\
Friction (static)                   & $0\ldots1008$                    & 1009\ldots2016                \\
Friction (dynamic)                    & $0\ldots1008$                    & 1009\ldots2016                \\
Flow stress (phip=0.1)            & $0\ldots4199$                    & 4200\ldots7799                \\
Battery 1 (10min)          & $0\ldots503$                     & 504\ldots634                  \\
Battery 2 (20min)          & $0\ldots1099$                    & 1100\ldots1637                \\
Nikuradse 1                     & $0\ldots229$                     & 230\ldots360                  \\
Nikuradse 2                     & $0\ldots199$                     & 200\ldots361                  \\
\end{tabular}
\label{train_test_table}
\end{table}


Models are compared based on their normalized mean of squared errors (NMSE),
\begin{equation}
\text{NMSE}(\hat y, y) = \frac{1}{\text{var}(y)} \text{MSE}(\hat y, y) = \frac{1}{\text{var}(y)} \sum_{i=1}^N \left(\hat y_i - y_i\right)^2,
\end{equation}
where var$(y)$ is the variance of the vector of target values $y$ and $\hat y$ is the vector of predictions from the model. The NMSE is in the range from zero to one and allows comparing regressors over multiple problem instances.

\subsection{Methods and Parameters}

\subsubsection{Operon and HeuristicLab}
Operon \cite{Burlacu2020} is an efficient state-of-the-art software implementation of genetic programming for symbolic regression. We use it here as a representative for symbolic regression systems based on genetic programming. Operon is implemented in modern C++ and relies heavily on thread-based parallelism to speed-up GP on multi-core machines. It stems out of the same research group that developed and maintains \emph{HeuristicLab} (HL) which is a software environment for heuristic and evolutionary algorithms implemented in C\# and uses the .NET platform \cite{Wagner2014}. It contains an implementation of tree-based genetic programming \cite{Kommenda2012} similar to Operon. We use both, Operon and HeuristicLab, with the same parameters to compare their relative performance.
The same parameters were used for all datasets.

\begin{table}
\caption{Parameters used for Operon and HeuristicLab.}
\centering
\begin{tabular}{ll}
Parameter & Value \\
\hline
Population size and generations & 1000, 100 \\
Selection & tournament with group size 5 \\
Mutation probability & 15~\% \\
Maximum tree length and depth & 100, 15 \\
Parameter optimization iterations & 10 \\
Function set & $+,\times,\div,\exp,\log,x^2,\sqrt{x},x^{1/3}$ \\
Loss function & mean of squared errors
\end{tabular}
\end{table}

\subsubsection{End-to-end Transformer:}
In the case of E2E \cite{DeepSR}, the authors had provided a web demo which provided easy inference for users based on a pretrained model that was provided. However, since the web server did not always respond and would sometimes give unexpected errors and also since web APIs in general can cause extra response time over what the true inference time would be, the inference was done locally using the pretrained checkpoint in a somewhat manual sense using the provided example Jupyter notebook after some modification.

For each dataset, the model checkpoint was loaded and then fit 20 times consecutively. For each instance, the model was run for inference on the training and test sets and the NMSE, R2 score, and training time were recorded. All runs were done on the RTX 2060 GPU except a few which were run on an M3 Pro chip, these exceptional runs will be highlighted in the results.

\subsubsection{SciMED:} 
This method combines evolutionary feature selection with GP-based automated machine learning (AutoML) for enriching the data domain and genetic programming for symbolic regression. Alternatively to GP-based symbolic regression SciMED provides computationally more expensive ``Las Vegas search SR'' for more stable and accurate results. In each phase it is possible to add domain knowledge. 

For our experiments, we used the library provided by the authors with default settings, and called the AutoML component (SciMED AutoML) and the GP-based SR component (SciMED GA-SR) separately. We did not use the evolutionary feature selection component or Las Vegas SR. We also did not use or test the capabilities of SciMED to add domain knowledge in each of the phases. It is important to note that the AutoML component simply calls TPOT~\cite{TPOT}, a GP-based AutoML package for Python. TPOT searches for the best pipeline including data preprocessing, feature selection methods, and regressors available. SciMED configures TPOT to use 5-fold cross-validation internally and refers to this step as the \emph{numerical} phase.
For the GA-SR component SciMED basically wraps the gplearn library and calls this the \emph{analytical} phase. Table~\ref{tab:scimed-parameters} lists the (default) parameters values for SciMED which we used for the AutoML and the GA-SR configurations.  
We executed 20 independent runs for each dataset, but found that the GA-SR phase produced the same results for all twenty iterations because the random seed of gplearn is not changed by SciMED. Training time was also recorded for all runs.

\begin{table}
\centering
\caption{Parameters used for SciMED AutoML and SciMED GA-SR.}\label{tab:scimed-parameters}
\begin{tabular}{lcc}
Parameter & SciMED AutoML & SciMED GA-SR  \\
\hline
Run times       &  1  & 20 \\
Generations     & 50  & 50\\
Population size & 100 & 100 \\
Parsimony coefficient & - & 0.05 \\
CV folds & 5 & 5 \\
\end{tabular}
\end{table}


\section{Results}

\begin{table}
\caption{NMSE values on training and testing sets as well as the runtime as observed in the experiments. The row with best NMSE value on the testing set for each dataset is marked in bold. Median and interquartile range over 30 independent runs are reported for Operon and HeuristicLab. Operon runtime is for 12 concurrent threads. 
The quality of Operon and HeuristicLab models is similar but Operon is approximately 8 times faster in single-core performance.}
\centering
\begin{tabular}{llccc}
Dataset & Software & NMSE (train) & NMSE (test) (rank) & Runtime [s] \\
\hline
Chemical Tower  & HeuristicLab       & 0.052           & 0.062 (3)          & 14487  \\
                & Operon             & 0.048           & 0.057 (2)          & 148 \\
                & E2E                & 52.34           & 55.99 (5)          & 351 \\
                & SciMED AutoML      & 0.000           & \textbf{0.025} (1)  & 12415 \\
                & SciMED GA-SR       & 0.512           & 0.525 (4)           & $\approx 600$ \\
\hline
Chemical Comp.       & HeuristicLab  & 0.092           & \textbf{0.204} (1) &  2975  \\
                     & Operon        & 0.092           & 0.270 (2)         &  29  \\
                     & E2E           & 0.774           & 1.229 (5)          & 92 \\
                     & SciMED AutoML & 0.028           & 0.448 (3)          & 7560 \\
                     & SciMED GA-SR  & 1.186           & 1.124 (4)          & $\approx 600$ \\
\hline
Flow Stress          & HeuristicLab  & 0.002           & 0.003 (2)         &  5425  \\
                     & Operon        & 0.000           & \textbf{0.001} (1) & 114   \\
                     & E2E\footnotemark[1]          & 0.422  & 0.491 (3) & 30 \\
                     & SciMED AutoML\footnotemark[3]   & 1.912   & 2.227 (4.5)  & 1198 \\
                     & SciMED GA-SR  & 1.912   & 2.227 (4.5) & $\approx 600$ \\
\hline
Friction (dyn.) & HeuristicLab         & 0.035           & \textbf{0.067} (1) &  3784 \\
                & Operon               & 0.047           & 0.070 (2)          &  18  \\
                & E2E                  & 1.087           & 1.087 (4)           & 229 \\
                & SciMED AutoML        & 0.062           & 0.261 (3)           & 1269 \\
                & SciMED GA-SR         & 332.2           & 453.3 (5)         & $\approx 600$\\
\hline
Friction (stat.) & HeuristicLab        & 0.065           & \textbf{0.095} (1)  &  3366  \\
                 & Operon              & 0.071           & 0.104 (2)          &  21 \\
                 & E2E              & 0.997           & 0.996 (4)           & 230 \\
                 & SciMED AutoML       & 0.050           & 0.202 (3)           & 6697 \\
                 & SciMED GA-SR        & 283.4           & 422.7 (5)         & $\approx 1200$ \\
\hline
Battery 1       & HeuristicLab         & 0.001           & \textbf{0.017} (1) &  2529 \\
                & Operon               & 0.000           & 0.024 (2)          &  18  \\
                & E2E                  & 0.058           & 0.347 (4)           & 79 \\
                & SciMED AutoML        & 0.003           & 0.051 (3)           & 1093 \\
                & SciMED GA-SR\footnotemark[4]           & N/A     & N/A    & $\approx 600$ \\
\hline
Battery 2       & HeuristicLab         & 0.001           & 0.152 (4)          &  3767 \\
                & Operon               & 0.001           & 0.100 (2)          &  35 \\
                & E2E                  & 0.175           & 0.151 (3)           & 171 \\
                & SciMED AutoML        & 0.003           & \textbf{0.035} (1)  & 1004 \\
                & SciMED GA-SR         & 0.575           & 0.684  (5)          & $\approx 600$ \\
\hline
Nikuradse 1     & HeuristicLab         & 0.001           & 0.056 (2)          &   578  \\
                & Operon               & 0.001           & \textbf{0.054} (1) &   5  \\
                & E2E                  & 0.304           & 0.905 (4)           & 35 \\
                & SciMED AutoML        & 0.001           & 0.734 (3)           & 716 \\
                & SciMED GA-SR         & 1.000           & 1.005 (5)           & $\approx 600$ \\
\hline
Nikuradse 2     & HeuristicLab         & 0.023           & \textbf{0.019} (1) &   282  \\
                & Operon               & 0.021           & 0.021 (3)          &   6  \\
                & E2E                  & 0.196           & 0.129 (4)           & 50 \\
                & SciMED AutoML        & 0.023           & 0.020 (2)           & 1778 \\
                & SciMED GA-SR         & 1.486           & 1.429 (5)           & $\approx 600$\footnotemark[5] \\
\end{tabular}
\label{results_table}
\end{table}
\footnotetext[1]{This was done using the web demo (1 run)}
\footnotetext[2]{M3 Pro runtime average}
\footnotetext[3]{Unanimous result across all 20 runs}
\footnotetext[4]{Would not run}
\footnotetext[5]{All SciMED GA-SR runs took about the same amount of time of around 10 minutes}

There are a few key points which can be seen across Table \ref{results_table}.
On average, Operon and HeuristicLab find the models with best NMSE on the testing partitions. Both implementations produce similar results but Operon is much faster (speedup $\approx$ 8) even after accounting for the fact that the HeuristicLab runtimes are for a single-threaded configuration while Operon used 12 concurrent threads. 

Only SciMED AutoML was able to produce models with comparable NMSE values on the testing partition. For two out of the nine datasets it even found the best models. On all other datasets either Operon or HeuristicLab produced better results. This is remarkable as SciMED AutoML uses TPOT internally which has access to the most important classical machine learning models such as random forests or extreme gradient boosting for trees (XGBoost). This again provides evidence that GP-based SR can produce models with an accuracy similar to more traditional black-box machine learning methods~\cite{affenzeller2020eurocast,la2021contemporary}.

Our results when using end-to-end transformers for symbolic regression (E2E) were much worse than the other methods. In some cases it even produced models with NMSE larger than one which is worse than a model predicting the mean of the target values. 
Operon had the best runtime on average and HeuristicLab had the worst runtimes. High dimensionality in the Chemical 1 and Chemical 2 datasets, and presence of categorical variables in the Friction (dyn.) and Friction (stat.) datasets influence model performance.
The statistical analysis in Table \ref{Average Rank} and Table \ref{Wilcoxon Tests} shows that on average HeuristicLab and Operon perform best over all datasets. Based on a Wilcoxon signed rank test, Operon and HeuristicLab are significantly better than E2E and SciMED GA-SR ($\alpha=0.05$). SciMED AutoML had medium performance.


\begin{table}
\centering
\caption{Statistical analysis of experiment results. }

\subfigure[Average ranks across datasets]{
\begin{tabular}{lc}
  Software & Average rank \\
  \hline
  HeuristicLab (HL)    & 1.8 \\
  Operon        (Op)   & 1.9 \\
  SciMED AutoML (ScML)   & 2.6 \\
  End-to-end Transformer (E2E)  & 4.0 \\
  SciMED GA+SR  (ScSR)  & 4.7  
\end{tabular}
\label{Average Rank}
}
\subfigure[Pairwise p-values]{
\begin{tabular}{l|ccccc}
       & Op   & ScML & E2E    & ScSR \\
\hline
  HL   & 0.91 & 0.25 & \textbf{0.004}  & \textbf{0.012} \\
  Op   &      & 0.30 & \textbf{0.008}  & \textbf{0.012} \\
  ScML &      &      & 0.055  & 0.096 \\
  E2E  &      &      &        & 0.570\\
  \\
\end{tabular}
\label{Wilcoxon Tests}
}
    \label{fig:combined}
\end{table}


\section{Discussion \& Conclusion}
In our experiments Operon remains the best all-around performer showing best to second best test errors and best to second best runtimes.
Even though we found many recent publications proposing new SR approaches based on deep learning in the literature review, in many cases no code was published or we did not succeed running the code. As an example AI~Descartes~\cite{Cornelio2023} requires a commercial solver (BARON) with yearly license costs of several hundred dollars even for a single seat academic license. Often code is published as academic abandonware together with a paper which does not run with up-to-date library versions. 
As a consequence, we only used the end-to-end transformer~\cite{Kim_2021} and the Scientist-Machine Equation Detector~\cite{keren2023computational} in our comparisons. However, the results of both systems do not reach the quality of results of tree-based genetic programming for symbolic regression as implemented in HeuristicLab or Operon.

\section*{Acknowledgements}
G.K. is supported by the Austrian Federal Ministry for
Climate Action, Environment, Energy, Mobility, Innovation and Technology, the
Federal Ministry for Labour and Economy, and the regional government of Upper Austria within the COMET project ProMetHeus (904919) supported by
the Austrian Research Promotion Agency (FFG).

\paragraph*{Author Contributions}
Y.R.: literature review, writing, running all experiments (except for Operon and HeuristicLab), data analysis. G.K.: conceptualization, preparation of datasets and running experiments for Operon and HeuristicLab. S.W.: conceptualization and oral presentation at the conference.

\bibliographystyle{splncs04}
\bibliography{bibliography}

\begin{thebibliography}{10}
\providecommand{\url}[1]{\texttt{#1}}
\providecommand{\urlprefix}{URL }
\providecommand{\doi}[1]{https://doi.org/#1}

\bibitem{affenzeller2020eurocast}
Affenzeller, M., Burlacu, B., Dorfer, V., Dorl, S., Halmerbauer, G., K{\"o}nigswieser, T., Kommenda, M., Vetter, J., Winkler, S.: White box vs. black box modeling: On the performance of deep learning, random forests, and symbolic regression in solving regression problems. In: Moreno-D{\'i}az, R., Pichler, F., Quesada-Arencibia, A. (eds.) Computer Aided Systems Theory -- EUROCAST 2019. pp. 288--295. Springer International Publishing, Cham (2020)

\bibitem{ESR}
Bartlett, D.J., Desmond, H., Ferreira, P.G.: Exhaustive symbolic regression. {IEEE} Transactions on Evolutionary Computation pp.~1--1 (2023). \doi{10.1109/tevc.2023.3280250}

\bibitem{biggio2020seq2seq}
Biggio, L., Bendinelli, T., Lucchi, A., Parascandolo, G.: A seq2seq approach to symbolic regression. Learning Meets Combinatorial Algorithms at NeurIPS2020  (2020)

\bibitem{biggio2021neural}
Biggio, L., Bendinelli, T., Neitz, A., Lucchi, A., Parascandolo, G.: Neural symbolic regression that scales. In: International Conference on Machine Learning. pp. 936--945. PMLR (2021)

\bibitem{Bomarito2021}
Bomarito, G., Townsend, T., Stewart, K., Esham, K., Emery, J., Hochhalter, J.: Development of interpretable, data-driven plasticity models with symbolic regression. Computers \& Structures  \textbf{252},  106557 (2021). \doi{10.1016/j.compstruc.2021.106557}

\bibitem{broløs2021approach}
Broløs, K.R., Machado, M.V., Cave, C., Kasak, J., Stentoft-Hansen, V., Batanero, V.G., Jelen, T., Wilstrup, C.: An approach to symbolic regression using feyn (2021)

\bibitem{Burlacu2020}
Burlacu, B., Kronberger, G., Kommenda, M.: Operon {C}++: {a}n efficient genetic programming framework for symbolic regression. In: Proceedings of the 2020 Genetic and Evolutionary Computation Conference Companion. pp. 1562--1570. GECCO '20, ACM (July 2020). \doi{10.1145/3377929.3398099}

\bibitem{Cornelio2023}
Cornelio, C., Dash, S., Austel, V., Josephson, T.R., Goncalves, J., Clarkson, K.L., Megiddo, N., El~Khadir, B., Horesh, L.: Combining data and theory for derivable scientific discovery with ai-descartes. Nature Communications  \textbf{14}(1), ~1777 (Apr 2023). \doi{10.1038/s41467-023-37236-y}

\bibitem{d2022deep}
d'Ascoli, S., Kamienny, P.A., Lample, G., Charton, F.: Deep symbolic regression for recurrent sequences. arXiv preprint arXiv:2201.04600  (2022)

\bibitem{dAscoli2024odeformer}
d'Ascoli, S., Becker, S., Mathis, A., Schwaller, P., Kilbertus, N.: {ODEF}ormer: Symbolic regression of dynamical systems with transformers. arXiv:2310.05573  (2023)

\bibitem{DEFRANCA2021609}
{de Franca}, F.O., {de Lima}, M.Z.: Interaction-transformation symbolic regression with extreme learning machine. Neurocomputing  \textbf{423},  609--619 (2021). \doi{10.1016/j.neucom.2020.10.062}

\bibitem{defranca2020interactiontransformation}
de~Franca, F.O., Aldeia, G.S.I.: {Interaction–Transformation Evolutionary Algorithm for Symbolic Regression}. Evolutionary Computation  \textbf{29}(3),  367--390 (09 2021). \doi{10.1162/evco\_a\_00285}

\bibitem{Goebel2008}
Goebel, K., Saha, B., Saxena, A., Celaya, J.R., Christophersen, J.P.: Prognostics in battery health management. IEEE Instrumentation \& Measurement Magazine  \textbf{11}(4),  33--40 (2008). \doi{10.1109/MIM.2008.4579269}

\bibitem{holt2023deep}
Holt, S., Qian, Z., van~der Schaar, M.: Deep generative symbolic regression. arXiv:2401.00282  (2023)

\bibitem{Kabliman2021}
Kabliman, E., Kolody, A.H., Kronsteiner, J., Kommenda, M., Kronberger, G.: Application of symbolic regression for constitutive modeling of plastic deformation. Applications in Engineering Science  \textbf{6},  100052 (June 2021). \doi{10.1016/j.apples.2021.100052}

\bibitem{DeepSR}
Kamienny, P.A., d'Ascoli, S., Lample, G., Charton, F.: End-to-end symbolic regression with transformers. arXiv preprint 2204.10532  (2022)

\bibitem{GE}
Kammerer, L., Kronberger, G., Burlacu, B., Winkler, S.M., Kommenda, M., Affenzeller, M.: Symbolic regression by exhaustive search: Reducing the search space using syntactical constraints and efficient semantic structure deduplication. In: Genetic Programming Theory and Practice {XVII}, pp. 79--99. Springer International Publishing (2020). \doi{10.1007/978-3-030-39958-0\_5}

\bibitem{keren2023computational}
Keren, L.S., Liberzon, A., Lazebnik, T.: A computational framework for physics-informed symbolic regression with straightforward integration of domain knowledge. Scientific Reports  \textbf{13}(1), ~1249 (2023)

\bibitem{Kim_2021}
Kim, S., Lu, P.Y., Mukherjee, S., Gilbert, M., Jing, L., Ceperic, V., Soljacic, M.: Integration of neural network-based symbolic regression in deep learning for scientific discovery. IEEE Transactions on Neural Networks and Learning Systems  \textbf{32}(9),  4166–4177 (Sep 2021). \doi{10.1109/tnnls.2020.3017010}

\bibitem{Kommenda2012}
Kommenda, M., Kronberger, G., Wagner, S., Winkler, S., Affenzeller, M.: On the architecture and implementation of tree-based genetic programming in heuristiclab. In: Proceedings of the 14th Annual Conference Companion on Genetic and Evolutionary Computation. p. 101–108. GECCO '12, Association for Computing Machinery, New York, NY, USA (2012). \doi{10.1145/2330784.2330801}

\bibitem{Kordon2008}
Kordon, A.: Evolutionary computation in the chemical industry. In: Yu, T., Davis, L., Baydar, C., Roy, R. (eds.) Evolutionary Computation in Practice, pp. 245--262. Springer Berlin Heidelberg (2008). \doi{10.1007/978-3-540-75771-9\_11}

\bibitem{Koza1992}
Koza, J.R.: Genetic Programming: On the Programming of Computers by Means of Natural Selection. MIT Press (1992)

\bibitem{Kronberger_2022}
Kronberger, G., de~Franca, F.O., Burlacu, B., Haider, C., Kommenda, M.: Shape-constrained symbolic regression—improving extrapolation with prior knowledge. Evolutionary Computation  \textbf{30}(1),  75–98 (2022). \doi{10.1162/evco\_a\_00294}

\bibitem{Kronberger2024}
Kronberger, G., Burlacu, B., Kommenda, M., Winkler, S.M., Affenzeller, M.: Symbolic Regression. CRC Press / Taylor Francis (2024)

\bibitem{Kronberger_2024_ppsn}
Kronberger, G., de~Franca, F.O., Desmond, H., Bartlett, D.J., Kammerer, L.: The inefficiency of genetic programming for symbolic regression. arxiv preprint 2404.17292  (2024)

\bibitem{Kronberger_GECCO_2018}
Kronberger, G., Kommenda, M., Promberger, A., Nickel, F.: Predicting friction system performance with symbolic regression and genetic programming with factor variables. In: Proceedings of the Genetic and Evolutionary Computation Conference. {ACM} (July 2018). \doi{10.1145/3205455.3205522}

\bibitem{Kubal_k_2020}
Kubalík, J., Derner, E., Babuška, R.: Symbolic regression driven by training data and prior knowledge. In: Proceedings of the 2020 Genetic and Evolutionary Computation Conference. GECCO ’20, ACM (Jun 2020). \doi{10.1145/3377930.3390152}

\bibitem{KUBALIK2021115210}
Kubalík, J., Derner, E., Babuška, R.: Multi-objective symbolic regression for physics-aware dynamic modeling. Expert Systems with Applications  \textbf{182},  115210 (2021). \doi{10.1016/j.eswa.2021.115210}

\bibitem{la2021contemporary}
La~Cava, W., Orzechowski, P., Burlacu, B., de~Fran{\c{c}}a, F.O., Virgolin, M., Jin, Y., Kommenda, M., Moore, J.H.: Contemporary symbolic regression methods and their relative performance. arXiv:2107.14351  (2021)

\bibitem{li2023metasymnet}
Li, Y., Li, W., Yu, L., Wu, M., Liu, J., Li, W., Hao, M., Wei, S., Deng, Y.: Metasymnet: A dynamic symbolic regression network capable of evolving into arbitrary formulations. arXiv preprint arXiv:2311.07326  (2023)

\bibitem{Makke2024}
Makke, N., Chawla, S.: Interpretable scientific discovery with symbolic regression: a review. Artificial Intelligence Review  \textbf{57}(1), ~2 (Jan 2024). \doi{10.1007/s10462-023-10622-0}

\bibitem{mundhenk2021symbolic}
Mundhenk, T.N., Landajuela, M., Glatt, R., Santiago, C.P., Faissol, D.M., Petersen, B.K.: Symbolic regression via neural-guided genetic programming population seeding. arXiv:2111.00053  (2021)

\bibitem{Nikuradse1933}
Nikuradse, J.: Laws of flow in rough pipes. Tech. rep., National Advisory Committee for Aeronautics Washington, NACA TM 1292 - Translation of "Strömungsgesetze in rauhen Rohren" VDI-Forschungsheft 361. Beilage zu “Forschung auf dem Gebiete des Ingenieurwesens" Ausgabe B Band 4, July/August 1933. (1950)

\bibitem{TPOT}
Olson, R.S., Bartley, N., Urbanowicz, R.J., Moore, J.H.: Evaluation of a tree-based pipeline optimization tool for automating data science. In: Proceedings of the Genetic and Evolutionary Computation Conference 2016. pp. 485--492. GECCO '16, ACM, New York, NY, USA (2016). \doi{10.1145/2908812.2908918}

\bibitem{petersen2021deep}
Petersen, B.K., Landajuela, M., Mundhenk, T.N., Santiago, C.P., Kim, S.K., Kim, J.T.: Deep symbolic regression: Recovering mathematical expressions from data via risk-seeking policy gradients (2021)

\bibitem{Reichardt2020}
Reichardt, I., Pallarès, J., Sales-Pardo, M., Guimerà, R.: Bayesian machine scientist to compare data collapses for the {N}ikuradse dataset. Physical Review Letters  \textbf{124}(8) (Feb 2020). \doi{10.1103/physrevlett.124.084503}

\bibitem{Saha2007}
Saha, B., Goebel, K.: Battery data set. Tech. rep., NASA Prognostics Data Repository, NASA Ames Research Center, Moffett Field, CA (2007), https://phm-datasets.s3.amazonaws.com/NASA/5.+Battery+Data+Set.zip, https://www.nasa.gov/content/prognostics-center-of-excellence-data-set-repository

\bibitem{Strathern1997}
Strathern, M.: ‘improving ratings’: audit in the british university system. European Review  \textbf{5}(3),  305–321 (Jul 1997). \doi{10.1002/(sici)1234-981x(199707)5:3<305::aid-euro184>3.0.co;2-4}

\bibitem{udrescu2020ai2}
Udrescu, S.M., Tan, A., Feng, J., Neto, O., Wu, T., Tegmark, M.: Ai feynman 2.0: Pareto-optimal symbolic regression exploiting graph modularity (2020)

\bibitem{udrescu2020ai}
Udrescu, S.M., Tegmark, M.: Ai feynman: a physics-inspired method for symbolic regression (2020)

\bibitem{Vaddireddy_2020}
Vaddireddy, H., Rasheed, A., Staples, A.E., San, O.: Feature engineering and symbolic regression methods for detecting hidden physics from sparse sensor observation data. Physics of Fluids  \textbf{32}(1) (Jan 2020). \doi{10.1063/1.5136351}

\bibitem{valipour2021symbolicgpt}
Valipour, M., You, B., Panju, M., Ghodsi, A.: Symbolicgpt: A generative transformer model for symbolic regression. arXiv:2106.14131  (2021)

\bibitem{vastl2022symformer}
Vastl, M., Kulh{\'a}nek, J., Kubal{\'\i}k, J., Derner, E., Babu{\v{s}}ka, R.: Symformer: End-to-end symbolic regression using transformer-based architecture. arXiv:2205.15764  (2022)

\bibitem{Vladislavleva2009}
Vladislavleva, E.J., Smits, G.F., {den Hertog}, D.: Order of nonlinearity as a complexity measure for models generated by symbolic regression via {P}areto genetic programming. IEEE Transactions on Evolutionary Computation  \textbf{13}(2),  333--349 (Apr 2009). \doi{10.1109/TEVC.2008.926486}

\bibitem{Wagner2014}
Wagner, S., Kronberger, G., Beham, A., Kommenda, M., Scheibenpflug, A., Pitzer, E., Vonolfen, S., Kofler, M., Winkler, S., Dorfer, V., Affenzeller, M.: Architecture and design of the {H}euristic{L}ab optimization environment. In: Klempous, R., Nikodem, J., Jacak, W., Chaczko, Z. (eds.) Advanced Methods and Applications in Computational Intelligence, Topics in Intelligent Engineering and Informatics, vol.~6, pp. 197--261. Springer (2014). \doi{10.1007/978-3-319-01436-4\_10}

\end{thebibliography}

\end{document}